# Evaluating COVID-19 Feature Contributions to Bitcoin Return Forecasting: Methodology Based on LightGBM and Genetic Optimization

Imen Mahmoud[1] and Andrei Velichko[2]

[1] University of Mannouba, Laboratory of Research in Innovative and Management, Risk, Accounting and Finance, Tunisia

*Email: Imen.mahmo@gmail.com

[2] Institute of Physics and Technology, Petrozavodsk State University, 185910 Petrozavodsk, Russia

*Email: Andrei Velichko velichkogf@gmail.com

ORCID:I.M. (0009-0006-3686-7249); A.V. (0000-0002-9341-1831)

### Abstract

This study proposes a novel methodological framework integrating a LightGBM regression model and genetic algorithm (GA) optimization to systematically evaluate the contribution of COVID-19-related indicators to Bitcoin return prediction. The primary objective was not merely to forecast Bitcoin returns but rather to determine whether including pandemic-related health data significantly enhances prediction accuracy. A comprehensive dataset comprising daily Bitcoin returns and COVID-19 metrics (vaccination rates, hospitalizations, testing statistics) was constructed. Predictive models, trained with and without COVID-19 features, were optimized using GA over 31 independent runs, allowing robust statistical assessment. Performance metrics ($R^2$, RMSE, MAE) were statistically compared through distribution overlaps and Mann–Whitney U tests. Permutation Feature Importance (PFI) analysis quantified individual feature contributions. Results indicate that COVID-19 indicators significantly improved model performance, particularly in capturing extreme market fluctuations ($R^2$ increased by 40%, RMSE decreased by 2%, both highly significant statistically). Among COVID-19 features, vaccination metrics, especially the 75th percentile of fully vaccinated individuals, emerged as dominant predictors. The proposed methodology extends existing financial analytics tools by incorporating public health signals, providing investors and policymakers with refined indicators to navigate market uncertainty during systemic crises.

### Acknowledgments

Special thanks to the editors of the journal and to the anonymous reviewers for their constructive criticism and improvement suggestions. This research was supported by the Russian Science Foundation (grant no. 22-11-00055-P, https://rscf.ru/en/project/22-11-00055/, accessed on 10 June 2025).

## 1. Introduction

The health disaster induced by the COVID-19 pandemic has had a significant impact on the financial sector, particularly on financial markets (Singh 2025), triggering unprecedented volatility and disrupting traditional asset classes (equities, commodities) (Ali et al. 2020; Zhang et al. 2020). This global catastrophe has accelerated interest in alternative assets such as Bitcoin and gold (Chemkha et al. 2021; Petti and Sergio 2024). The study by Mselmi and Mahmoud (2024) reveals that cryptocurrencies, particularly Bitcoin, have partially replaced gold as safe haven assets during the pandemic, highlighting their growing role in hedging strategies in times of crisis. Bitcoin, as a cryptocurrency, embodies a new form of decentralized and digital money and has revolutionized financial markets through blockchain technology, competing with traditional currencies and precious metals as a means of preserving value. As a financial asset not directly linked to the real economy, Bitcoin remains an excellent hedge against the risks associated with products influenced by the macroeconomic market (Luo 2020). In response to this volatility, researchers from several disciplines have undertaken to examine the impacts of the pandemic. Research has primarily focused on the relationships between the pandemic and asset prices such as commodities (especially energy like crude oil and natural gas), precious metals (gold and silver), and cryptocurrencies, including Bitcoin, which form the foundation of the global economy (Niamkova and Moreira 2023). Machine learning (ML) has become a critical tool for modeling these complex market dynamics. With the growing importance of this topic, various methodologies, including LSTM and neural networks, have been used to forecast financial asset prices using machine learning algorithms. However, researchers still consider the study of financial asset price prediction during pandemic shocks to be an emerging field. Most studies have attempted to predict asset prices using machine learning, but few have incorporated COVID-19-related features.

This is the approach we use to examine how machine learning methods can help predict Bitcoin price performance during times of economic disruption, including pandemic-related factors. How can incorporating COVID-19-related indicators improve the accuracy of machine learning models in predicting Bitcoin returns during pandemic-induced uncertainty and volatility? This study highlights the potential of machine learning techniques (Velichko et al. 2022), combined with a set of pandemic-related indicators, to reflect the impact of COVID-19 and improve Bitcoin value prediction. This study is important because it introduces an innovative methodological framework specifically designed to quantify and statistically validate the influence of pandemic-related indicators on Bitcoin returns during periods of significant market disruption. Unlike previous studies, our approach systematically integrates health data such as vaccination rates, hospitalization statistics, and testing metrics into machine learning modeling, significantly enhancing predictive accuracy for extreme price fluctuations.

In practical terms, the proposed methodology allows policymakers to develop targeted financial stability measures informed by robust pandemic-market linkages. Additionally, investors can more effectively hedge portfolios against systemic risks through refined signals derived from health data. Methodologically, the incorporation of permutation feature importance (PFI) analysis within a genetically optimized LightGBM framework represents a meaningful advancement in the existing financial forecasting literature, clearly quantifying feature importance during unstable market conditions. Lastly, our findings also contribute to the ongoing debate regarding market efficiency, providing empirical evidence supporting the adaptive market hypothesis during crisis periods.

## 2. Literature review

2.1 Machine Learning Models for Forecasting

Recent implementations of machine learning have garnered heightened interest, particularly in the realm of cryptocurrency price forecasts like Bitcoin (Derbentsev, Matviychuk, et al. 2020). Conventional models, including linear regression and support vector machines, have progressively been supplanted by more sophisticated architectures, such as long short-term memory networks (Jiménez-Carrión and Flores-Fernandez 2025; Rodrigues and Machado 2025) and Random Forest, XGBoost, GBM (Ensembles), which are more adept at handling complex time series data (Derbentsev, Datsenko, et al. 2020; Derbentsev et al. 2021; Qureshi et al. 2025). A multitude of studies has examined various forecasting models, Boozary et al. (2025) reviewed studies on machine learning (ML) specifically tailored to Bitcoin price prediction, focusing on evaluating the robustness, accuracy, and relevance of advanced machine learning techniques such as long-short-term memory (LSTM) networks.

Rathore et al. (2022) showed that LSTM models work well for short-term predictions because they can store sequential data. McNally et al. (2018) also compared several ML approaches and found that recurrent neural networks (RNNs) are more effective than classical methods. Patel et al. (2015) combined several ML approaches, such as SVM, ANN, and random forest, to make them more robust. Freeda et al. (2021) also looked at hybrid models and found that ensemble methods work better than individual ronments like the cryptocurrency market.

Greaves and Au (2015) investigated the application of machine learning to forecast future fluctuations in Bitcoin prices by leveraging blockchain-derived characteristics. Their methodology attains a classification accuracy of approximately 55% in predicting the fluctuation of Bitcoin prices. Jiang (2020) investigated the short-term price performance of Bitcoin using machine learning techniques with an emphasis on hourly data rather than daily data. LSTM and GRU models demonstrated superior performance over MLP, providing improved short-term accuracy and facilitating real-time trading opportunities. K. Ibrahim and Singh (2025) investigated the application of machine learning models to forecast Bitcoin prices utilizing real-time historical data. The authors employed linear regression, support vector machines (SVMs), and LSTM recurrent neural networks to examine Bitcoin price volatility. The findings indicate that deep learning methodologies, such as LSTMs, are proficient in identifying the intricate and fluctuating patterns within the bitcoin market. Similarly, Omole and Enke (2024) tried to predict the price movements of Bitcoin and evaluate the profitability of trading methods based on these predictions, using the CNN-LSTM model. Their results indicate that with Boruta features, the model achieves an accuracy of 82.44%, thus generating significant returns in trading strategies

These researches, despite their diverse methodologies and objectives, converge on a common observation: machine learning and deep learning techniques, in particular sequential models such as LSTM, have become indispensable for predicting financial asset price variations. Nevertheless, despite considerable progress, many challenges remain, including the need for more resilient models to handle the extreme volatility of cryptocurrencies, the incorporation of alternative data (such as social media and technical data), and the expansion of analyses to encompass longer times and a more diverse range of assets. The 2019 health crisis, for example, was marked by a high degree of volatility and major economic shocks. This crisis highlighted not only the observed change in the functioning of financial markets but also the need to develop decision support systems to understand these external events in a context of uncertainty.

## 2.2 Impact of COVID-19 on Forecasting

The COVID-19 outbreak has caused significant disruptions in financial markets, with stock indices falling and prices of precious metals and commodities, including cryptocurrencies, becoming more available. Many studies have focused on the impact of pandemic-related variables on predicting financial asset prices (Lorenz et al. 2022). They use health status data (confirmed cases, hospitalizations, testing, infection rates) and use user sentiment on social media about the pandemic. For exemple Luo (2020) uses 4 machine learning models to predict Bitcoin price performance and trajectory from Twitter, combined with COVID-19 statistics (cases, recoveries, deaths) in the 7 countries where the majority of Bitcoin holders are located. The study found that social data significantly improved predictions, while health data had minimal impact during the examined period. Apergis (2022), on the other hand, found that adding COVID-19-related factors improves the accuracy of volatility predictions for cryptocurrency returns, such as Bitcoin, Ethereum, and Litecoin, by using two metrics: the total number of confirmed cases around the world and the death toll. Niamkova and Moreira (2023) confirmed the premise that the COVID-19 pandemic has heightened the demand for cryptocurrencies and that COVID-19-related data can enhance the forecast of Bitcoin's price through two models: Random Forest and Long Short-Term Memory.

Chatterjee et al. (2024) attempted to predict the price of Bitcoin by combining COVID-related variables using time series models (ARIMA, VAR, and VECM) and machine learning models (Random Forest, LSTM, and GRU) before and during the COVID-19 pandemic. Their results indicate that COVID-19 data are significant predictors of Bitcoin price during the pandemic. Bitcoin has served as a safe haven amid this crisis. The pandemic's impact has also influenced the prices of precious metals, including gold, beyond cryptocurrency. Mohtasham Khani et al. (2021) used pandemic data, such as COVID-19 case statistics (new and total cases), along with LSTM models to improve the accuracy of gold price predictions.

Recent research shows that Bitcoin-based forecasting methods don't work well during emergencies. For instance, Jiang (2020) and Ibrahim & Singh (2022) have both talked about how well LSTM/CNN models work for predicting Bitcoin prices. These studies, on the other hand, don't take into account two important aspects of the COVID-19 crisis: the systematic inclusion of COVID-19 indicators as technical attributes, like vaccination rates and ICU admissions (Luo 2020; Niamkova and Moreira 2023), which show behaviors that conventional models don't fully capture, and the changing nature of market efficiency during crises, as shown in the study by (Mahmoud 2025).

All these shortcomings prevent conventional models from capturing the nonlinear behaviors induced by pandemic shocks. Although market efficiency remains a matter of debate (Htun et al., 2024), the lack of clear quantitative correlations between Bitcoin volatility and public health data (Foroutan and Lahmiri 2024) highlights a deeper methodological problem: current approaches either ignore the structuring variables of the pandemic or treat them inadequately (e.g., vaccination rates as raw data rather than as statistical transformations). Short-term bias occurs when excessive attention is paid to very short periods, such as hours or days, thereby ignoring larger changes caused by the pandemic, which last longer than 7 days.

Numerous studies have demonstrated the performance of machine learning models for cryptocurrency forecasting and identified the impact of COVID-19 indicators on the markets. However, some gaps remain, including the absence of systematic integration of granular health indicators (such as vaccination rates and intensive care admissions), a lack of rigorous methods to quantify their predictive importance, and the fact that most studies focus on the short term at the expense of understanding prolonged crisis dynamics. Our research addresses these limitations through an innovative approach by combining:

- A LightGBM model optimized by a genetic algorithm
- We integrated 45 pandemic indicators that were statistically transformed using sliding windows.
- A permutation importance analysis (PFI) is conducted to scientifically assess the contribution of the pandemic indicators.
- The forecast horizon is set to 7 days and is specifically adapted to address prolonged economic shocks.

This method differs from more conventional LSTM/CNN approaches based on the optimal compromise made not only on the optimized hyperparameters during variable selection but also on the relevant use of the indicators relied on to preliminarily challenge the originality of a discrete model ($R^2$ and RMSE for extreme variations and MAE for central behaviors), which highlight how choosing the right model alone is insufficient to explain the exceptional market behaviors during periods of health crises.

## 3. Methodology

The study quantifies how much COVID-19 information improves short-term Bitcoin return forecasts. To isolate that effect we run two otherwise-identical modelling pipelines:

1) Baseline configuration – uses only
    a) Lagged Bitcoin log-returns
    b) Cyclic calendar variables (DayOfWeek_cos, DOY_cos)
2) COVID-augmented configuration – uses everything in the Baseline plus
    a) 45 daily COVID-19 metrics (case counts, hospital-load indicators, vaccination progress, testing statistics, policy stringency, etc.)

The key design choices are:

1) Forecasting task – predict the log-return seven days ahead.
2) Feature window – for each date $t$ we summarise the previous 14 days ($t − 14 \ldots t$) with window statistics (mean, median, percentiles, range, last-first difference, etc.) applied to every predictor.
3) Learner – LightGBM regression; hyper-parameters are tuned with a genetic algorithm (150 generations) that simultaneously selects the most informative feature subset.
4) Evaluation split – chronological train/test cut: first 358 observations for training, last 200 for testing (Dec 11 2020 → June 21 2022, inclusive).
5) Robustness – the GA optimisation is repeated 31 independent times; we compare the resulting distributions of $R^2$, MAE and RMSE for the Baseline and COVID-augmented models with a two-sided Mann-Whitney U test.

If the COVID-augmented runs show a statistically significant uplift in $R^2$ (and/or corresponding error reduction), we conclude that pandemic indicators carry exploitable information for seven-day Bitcoin return forecasts.

### 3.1. Data Description

The study is based on data gathered from Yahoo Finance regarding the closing daily prices of Bitcoin between December 11, 2020, and June 21, 2022. To standardize price fluctuations observed during this period, which was marked by significant volatility related to the COVID-19 pandemic and post-pandemic market fluctuations, daily returns were converted to logarithms.

*Auxiliary Temporal Features*: Two cyclic temporal features were computed:
- DayOfWeek_cos: Cosine transformation of the day of the week to capture weekly seasonality.
- DOY_cos: Cosine transformation of the day of the year (DOY) to capture annual seasonality.

*COVID-19 Features Dataset*: An extensive set of pandemic-related indicators were included to reflect the impact of COVID-19 on asset returns:

- Total cases, new cases smoothed, total deaths, new deaths smoothed.
- Metrics normalized per million inhabitants (e.g., total cases per million).
- Healthcare system load indicators (ICU patients, hospital admissions).
- Vaccination data (total vaccinations, fully vaccinated individuals, boosters administered).
- Testing statistics (total tests, positive rates).
- Policy stringency index.

The dataset contained daily observations of Bitcoin returns, totaling 558 data points after ensuring temporal alignment. For predictive modeling, the target variable was defined as the Bitcoin return shifted by 7 days forward. Thus, the model was trained to forecast Bitcoin returns 7 days into the future using historical data within a maximum rolling window of 14 preceding days. Two modeling approaches were examined: one utilizing exclusively historical Bitcoin returns, and another enhanced by incorporating additional COVID-19-related features as described earlier.

For model evaluation, the initial 358 observations were used as the training set, and the remaining 200 observations served as the test set.

**3.2 LightGBM Model Parameter Optimization**

Predictive modeling was carried out using the LightGBM (Light Gradient Boosting Machine) regression algorithm available via the lightgbm Python library. Model hyperparameters were optimized through genetic algorithm techniques implemented using the pygad Python library. The genetic algorithm systematically explored hyperparameter spaces to maximize predictive performance (measured primarily via the $R^2$ metric).

Optimized hyperparameters for the LightGBM model included:

- boosting_type: ["gbdt", "dart", "goss"]
- num_leaves: Integer range [1, 100]
- max_depth: [-1, 5, 10, 15, 20]
- learning_rate: Continuous range [0.001, 0.1]
- n_estimators: Integer range [3, 200]
- subsample: Continuous range [0.5, 1.0]
- colsample_bytree: Continuous range [0.5, 1.0]
- min_child_samples: Integer range [10, 50]
- reg_alpha: Continuous range [0.0, 1.0]

- reg_lambda: Continuous range [0.0, 1.0]

**3.3. Feature Engineering and Window-Based Optimization**

Additional feature extraction was performed through temporal window analysis, which utilized historical data with a look-back window ranging from 1 to 14 days. In this setup, $w0\_i$ represents the start offset of the window relative to the current date (ranging from 0 to 8, where positive values refer to steps back in time), and $wl\_i$ defines the window length (ranging from 1 to 7 days). For each window, a range of statistical and entropy-based features was computed (Table 1).

**Table 1** Feature Extraction Function Codes and Abbreviations. The following table lists the function codes (*FC*) used during temporal window-based feature extraction, along with their corresponding descriptions and abbreviated suffixes used in column naming conventions:

| Function Code (*FC*) | Description | Abbreviation Suffix |
|---|---|---|
| -1 | Empty column (no features extracted) | - |
| 0 | Raw window values | _{j+1} (e.g., _1, _2) |
| 1 | Window mean (average) | _avg |
| 3 | Window median | _median |
| 4 | Maximum value in window | _max |
| 5 | Minimum value in window | _min |
| 6 | Range of values in window (max-min) | _range |
| 7 | Sum of window values | _sum |
| 8 | First value in window | _first |
| 9 | Last value in window | _last |
| 10 | Difference (last - first) | _diff |
| 11 | Percentage change (relative change) | _pctch |
| 12 | 25th percentile | _p25 |
| 13 | 50th percentile | _p50 |
| 14 | 75th percentile | _p75 |
| 15 | Interquartile range | _IQR |

The function code $FC = 2$ was reserved for specific entropy-based functions, which were not utilized in this particular study.

Through the described methodology, the optimal combination of features and model parameters was systematically identified to maximize predictive accuracy for financial asset returns.

**3.4. LightGBM Performance Metrics and Permutation Feature Importance (PFI) Analysis**

The predictive ability of the LightGBM model was evaluated with three complementary metrics calculated for each of the 31 experimental cases:

1. $R^2$ (coefficient of determination): proportion of variance in Bitcoin returns explained by the model ($0 \leq R^2 \leq 1$; higher values denote better fit).

2. MAE (mean absolute error): average absolute difference between predicted and actual returns; lower values indicate higher accuracy.
3. RMSE (root-mean-square error): square-root of the mean squared prediction error; like MAE, lower values signify better performance but penalise larger errors more strongly.

After establishing baseline performance, we quantified the contribution of every input variable using Permutation Feature Importance (PFI) with respect to $R^2$. In the PFI procedure each feature is randomly shuffled while all others are left unchanged, and the consequent drop in $R^2$ is recorded. The size of this decrease shows how strongly that feature influences model accuracy: the larger the reduction, the more essential the feature. Ranking features by their PFI scores therefore yields a transparent, quantitative ordering of variable importance.

This combined evaluation—first benchmarking overall predictive quality via $R^2$, MAE, and RMSE, then dissecting individual contributions with PFI—enabled us to identify the most influential predictors and to focus subsequent modelling efforts on variables that demonstrably enhance forecast performance.

**3.5. Methodology for Evaluating the Impact of COVID-19 Features on Bitcoin Return Prediction**

To assess the impact of COVID-19-related features on the predictive accuracy of Bitcoin returns, we implemented a comprehensive methodological approach combining machine learning and statistical analysis (Figure 1).

First, a prediction module based on the LightGBM algorithm was established, including training and testing stages. This module received as input a set of hyperparameters described in Section 3.2 and a combination of features constructed according to the procedures detailed in Section 3.3 ("Feature Engineering and Window-Based Optimization"). Specifically, this module trained the LightGBM model on a predefined training dataset, and subsequently, its predictive performance was evaluated on a separate test dataset. The model's predictive accuracy was primarily quantified using the $R^2$ metric, which measures how well the predicted values fit the actual observed values in the test set. Additionally, two complementary metrics—Mean Absolute Error (MAE) and Root Mean Square Error (RMSE)—were computed to further evaluate predictive accuracy.

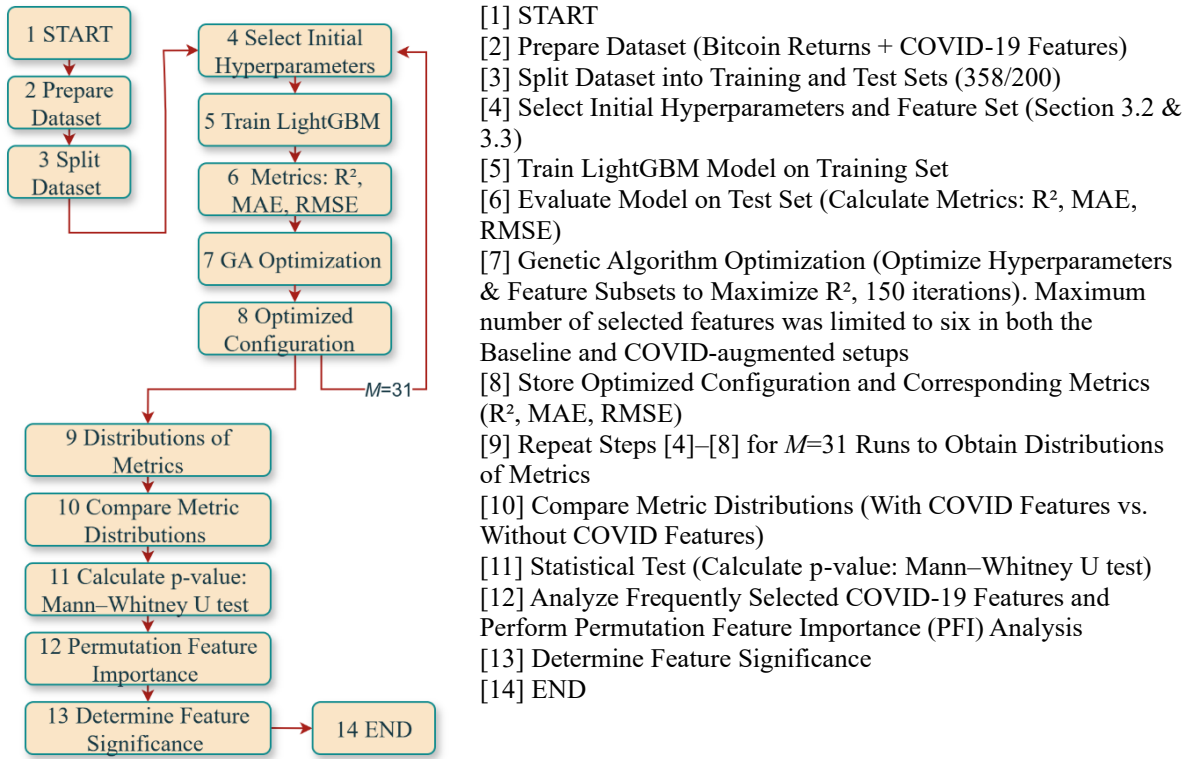

**Figure 1.** Block diagram of the proposed methodology for evaluating the impact of COVID-19-related features on Bitcoin return prediction using LightGBM modeling and genetic algorithm optimization.

Next, this entire prediction module (training and evaluation) was embedded into a genetic algorithm (GA) optimization framework, aimed at maximizing the R² metric. The genetic algorithm iteratively optimized both the hyperparameters of the LightGBM model and the subset of input features over 150 iterations per optimization run.

To ensure a fair and consistent basis for comparison between the two configurations, the maximum number of selected features was limited to six in both the Baseline and COVID-augmented setups. This constraint enforced equal model complexity, allowing us to isolate and assess the contribution of COVID-19 features under comparable predictive conditions.

Thus, at the conclusion of each GA optimization process, we obtained an optimally configured predictive model, characterized by its corresponding optimal feature subset and tuned hyperparameters. It is important to note that the stochastic nature of the genetic algorithm implies that different optimization runs may yield varying optimal configurations of hyperparameters and feature subsets.

To ensure robustness and statistically sound evaluation, we repeated this GA optimization process independently for $M$ runs. In this study, $M$ was set to 31. Consequently, we generated a distribution of optimal metric values (R², MAE, RMSE) over these multiple optimization runs, separately for two distinct cases: (1) models trained without COVID-19-related features, and (2) models trained with COVID-19-related features included.

Following these runs, histograms were constructed to visualize the distributions of the obtained performance metrics, allowing direct comparison between scenarios "without COVID" and "with COVID". Specifically, the degree of overlap between histograms was quantified statistically to determine the significance of observed differences in metric distributions. A formal statistical hypothesis test (Mann–Whitney U test) was applied, enabling us to calculate p-values that quantified whether the observed differences in predictive performance between the two scenarios were statistically significant.

Finally, beyond comparing aggregated performance metrics, we analyzed which COVID-19-related features were most frequently selected by the genetic algorithm across multiple optimization runs. By identifying these commonly selected features and combining these findings with the independently obtained Permutation Feature Importance (PFI) analysis described previously (Section 3.4), we gained deeper insight into the individual importance of COVID-19-related predictors and their relative contribution to forecasting Bitcoin returns.

This rigorous, statistically validated, and iterative methodological approach provided strong evidence regarding whether and to what extent the inclusion of COVID-19-related features enhanced the predictive performance of the LightGBM model in forecasting Bitcoin returns.

**4. Results and Discussion**

**4.1. Impact of COVID-related features on Bitcoin price**

Figure 2 presents the histograms illustrating the distributions of performance metrics obtained for models with and without COVID-related features. The analysis aimed to evaluate the impact of COVID-related features on Bitcoin price prediction by systematically comparing predictive models constructed with versus without COVID data. The significance of observed differences was assessed statistically, resulting in the following key findings:

- $R^2$: The mean $R^2$ value notably increased from 0.091 (no COVID) to 0.128 (with COVID, +40%), reflecting a statistically significant improvement (p-value = $2.95 \cdot 10^{-9}$).
- MAE: Mean Absolute Error decreased slightly from 0.03428 (no COVID) to 0.03409 (with COVID). However, the improvement was marginal and not statistically significant (p-value = 0.0929), indicating a less clear impact.
- RMSE: Root Mean Square Error demonstrated clear improvement, with the mean RMSE dropping from 0.04586 (no COVID) to 0.04492 (with COVID, -2%), which was highly significant statistically (p-value = $3.36 \cdot 10^{-9}$).

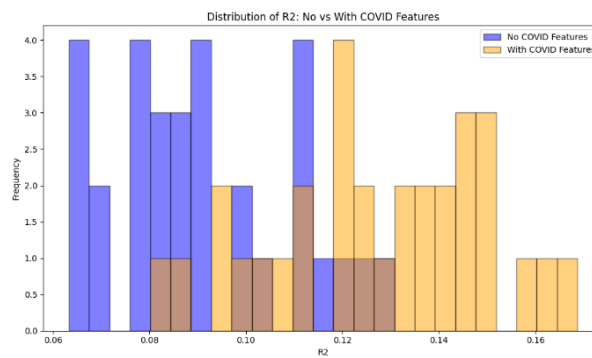

(a)

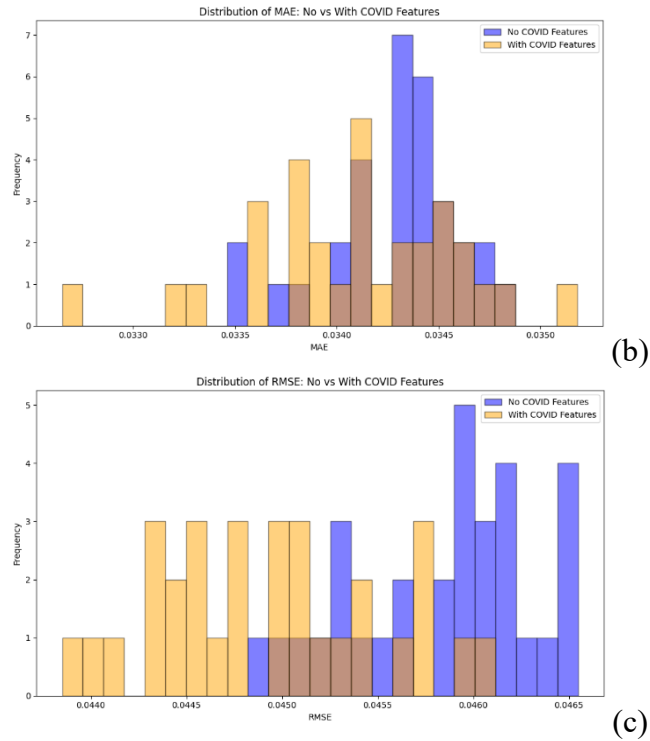

**Figure 2.** Comparison of metric distributions between models with and without COVID-related features: (a) $R^2$ (coefficient of determination), clearly illustrating improved variance explained when COVID features are included; (b) MAE (Mean Absolute Error), showing substantial overlap indicating minimal impact; and (c) RMSE (Root Mean Square Error), demonstrating notable error reduction in models incorporating COVID features.

Histograms illustrating metric distributions provide further insights (Figure 2):

For $R^2$ and RMSE, distributions distinctly separate the "with COVID" scenario from the "no COVID" scenario, clearly visualizing the positive influence of COVID-related features.

For MAE, however, the distributions overlap substantially, aligning with the non-significant statistical test, confirming that COVID features had minimal impact according to this metric.

The analysis thus confirms that COVID-related indicators significantly enhance Bitcoin price prediction in terms of variance explained ($R^2$) and overall error reduction (RMSE), whereas improvements in average prediction accuracy (MAE) are less evident.

By contrast, MAE being less influenced by outliers indicates that daily predictions remain generally steady, whether or not those variables are included.

The $R^2$ and RMSE results suggest that pandemic-related factors have a constructive influence. These metrics help the model better capture the complexities of market behavior during unstable periods, such as sudden downturns or spikes (Figure 3) by more accurately reflecting variations ($R^2$) and reducing general prediction errors (RMSE).

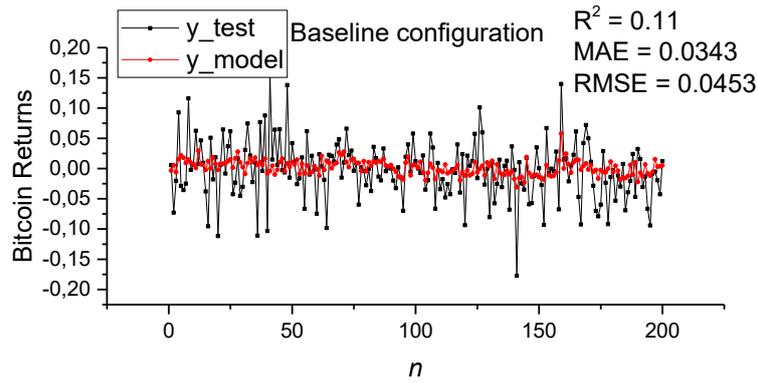

a)

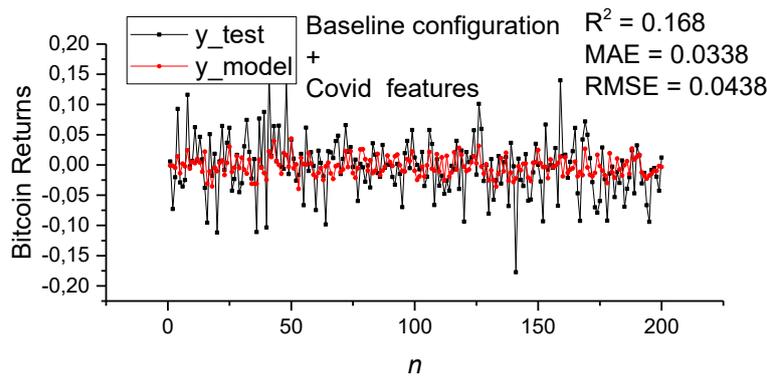

b)

**Figure 3**. Predicted (y_model) vs actual Bitcoin (y_test) returns for the test set: a) Baseline configuration, using only: Lagged Bitcoin log-returns and Cyclic calendar variables (DayOfWeek_cos, DOY_cos) b) COVID-augmented configuration, which includes the baseline features plus 45 COVID-related metrics (case counts, hospital load, vaccination progress, etc.).

The results in Figure 3 further illustrate the impact of COVID-related features. The $R^2$ value increases from 0.11 to 0.168 when COVID features are included, indicating a higher proportion of variance explained. Visually, the red predicted line in the COVID-augmented configuration more frequently captures the direction of spikes and dips in the Bitcoin returns. The RMSE also shows slight improvement, while the MAE remains similar — in line with previous statistical observations. These findings reinforce that COVID-related metrics help the model better adapt to short-term market dynamics, particularly during turbulent periods.

However, the MAE values tests showing no significant improvement. This implies that COVID-related variables do not enhance the model's ability to predict average fluctuations in review scores. Instead, they seem to be more helpful when it comes to modeling unusual or extreme cases, which have less impact on the MAE. This difference in how each metric responds helps explain the divergence between RMSE and MAE. RMSE is more responsive to larger errors, highlighting improvements in capturing extreme changes tied to COVID variables.

Further investigation into specific COVID-related features that most strongly contributed to these improvements is described in the following section.

## 4.2 Analysis of COVID-19 Related Features and Their Statistical Significance in Forecasting Bitcoin Returns

In Figures 4,5, the feature importance determined by PFI analysis is presented for two scenarios corresponding to the highest $R^2$ values obtained in our experiments: (Figure 4) without COVID-related features (maximum $R^2 = 0.13$), and (Figure 5) including COVID-related features (maximum $R^2 = 0.168$). Feature importance was quantified using the $R^2\_PFI$ metric, which indicates the reduction in model accuracy ($R^2$) when a given feature is randomly shuffled while keeping all other features intact. Larger values of $R^2\_PFI$ thus imply greater feature influence on predictive performance.

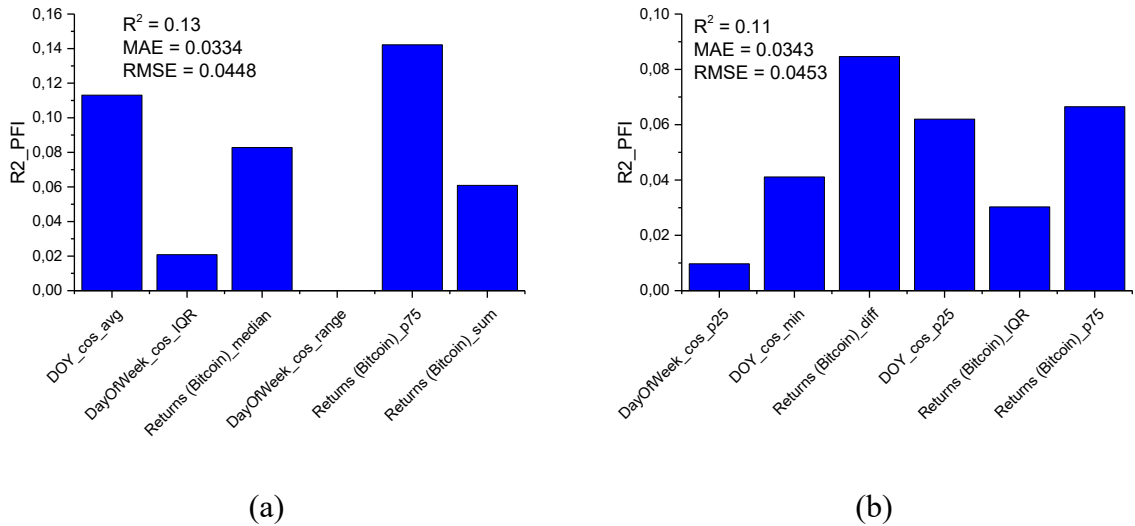

(a)               (b)

**Figure 4**. The PFI feature importance for scenarios with maximum $R^2$ values without COVID-related features (maximum $R^2 = 0.13$) and (b) case $R^2 = 0.11$ The feature importance is quantified by the R2_PFI value, indicating the degree of influence each feature has on model performance.

In the first scenario (Figure 4a), where no COVID-related features were included, the most influential feature was Returns (Bitcoin)_p75 ($R^2\_PFI = 0.1422$), representing patterns captured by the 75th percentile of Bitcoin returns within the historical 7-day window, reflecting the importance of extreme return behaviors. This was closely followed by DOY_cos_avg ($R^2\_PFI = 0.1131$), an auxiliary temporal feature capturing average annual seasonality effects. Conversely, the feature DayOfWeek_cos_range, representing weekly cyclical variability within the 7-day window, exhibited negligible influence ($R^2\_PFI = 0.0000$). In the second randomly selected configuration after GA optimization (Figure 4b, $R^2 = 0.11$), the most informative predictors remained those derived from Bitcoin return dynamics and cyclic temporal variables. Specifically, the feature Returns (Bitcoin)_diff demonstrated the highest contribution to the model ($R^2\_PFI \approx 0.085$). This feature corresponds to the difference between the last and first values of the Bitcoin return series within the historical 7-day rolling window (see Table 1, Function Code 10). Its importance suggests that short-term directional trends—whether increasing or decreasing—within recent return intervals offer substantial predictive power.

Other notably contributing features included Returns (Bitcoin)_p75 (the 75th percentile of returns, capturing upward spikes) and DOY_cos_p25 (capturing annual periodic behavior around the lower seasonal amplitude), reinforcing the relevance of both asymmetric return distributions and seasonality effects. Conversely, the least important was DayOfWeek_cos_p25, again pointing to the limited role of intra-week patterns in this configuration.

In the second scenario (Figure 5a), when COVID-related features were included, people_fully_vaccinated_p75 (75th percentile of fully vaccinated individuals per million inhabitants over a 7-day window) demonstrated the highest importance ($R^2\_PFI = 0.1174$). This finding underscores the substantial impact of critical vaccination thresholds on market behavior and predictive accuracy. Another highly significant predictor was hosp_patients_p75 (75th percentile of hospitalized COVID-19 patients per million), reflecting pressures on healthcare systems as strong market stress signals, followed by new_people_vaccinated_smoothed_p50 (median of daily smoothed counts of newly vaccinated individuals), both exhibiting notable influence ($R^2\_PFI \approx 0.075$). Thus, inclusion of pandemic-related indicators significantly alters the relative importance of input features, emphasizing their critical role in improving the accuracy of Bitcoin return forecasts. Figures 5a and 5b depict the most influential features when COVID-related variables are included in the model. In Figure 5a, it is particularly noteworthy that no Bitcoin return-based features appear among the top contributors. The model achieves an $R^2$ of 0.168 solely through pandemic-related variables, suggesting that under certain configurations, epidemiological dynamics alone can effectively predict short-term market behavior. This highlights the dominant role of vaccination progress and healthcare stress signals (e.g., hospitalizations) in shaping financial expectations during crisis periods.

By contrast, Figure 5b corresponds to a scenario where genetic algorithm (GA) optimization proposed a hybrid input set. Here, Returns (Bitcoin)_p75 re-emerges as a significant feature, on par with the top COVID predictors such as weekly_hosp_admissions_IQR and new_vaccinations_smoothed_IQR. This configuration achieves an $R^2$ of 0.125, indicating that a combination of behavioral financial features and pandemic indicators can also yield effective predictions.

These two contrasting cases further support the rationale of our proposed approach: multiple equally valid feature combinations can successfully model the same target series. Analyzing the distribution of feature importance across such configurations offers deeper insights into the mechanisms underlying model performance and the dynamic relevance of various predictors.

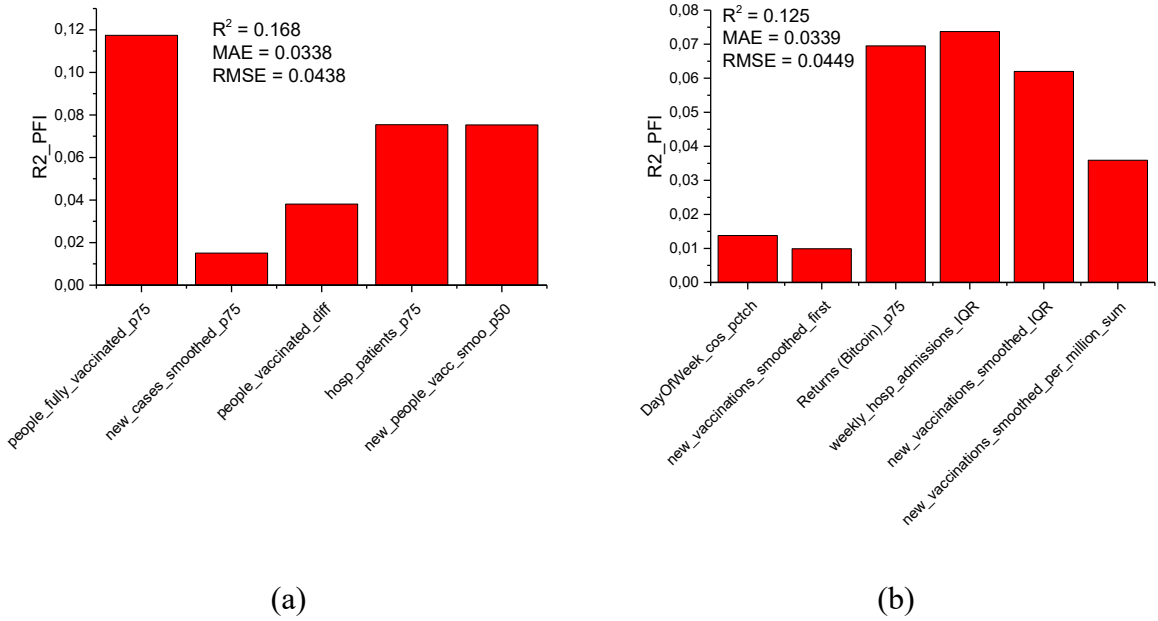

(a)                              (b)

**Figure 5**. The PFI feature importance for scenarios with maximum R² values including COVID-related features (maximum R² = 0. 168) and (b) case R² = 0.125. The feature importance is quantified by the R2_PFI value, indicating the degree of influence each feature has on model performance.

Further statistical analysis of the frequency of occurrence (denoted as "frequency," the number of times a feature appeared in the optimal feature subsets selected by the genetic algorithm across multiple modeling runs) in models built without COVID-related features revealed the following patterns:

First-Level Features (most influential):

- Returns (Bitcoin)_p75 (75th percentile of returns, average R²_PFI = 0.106; frequency = 6): Indicates consistent relevance of capturing upper-tail behavior in Bitcoin returns.

- Returns (Bitcoin)_p25 (25th percentile of returns, average R²_PFI = 0.051; frequency = 4): Emphasizes the predictive value of lower-tail return behaviors.

- Returns (Bitcoin)_avg (average return, average R²_PFI = 0.056; frequency = 3) and Returns (Bitcoin)_pctch (percentage change from first to last day, average R²_PFI = 0.064; frequency = 3): Reflect the importance of overall return levels and short-term directional trends.

Second-Level Features:

- Returns (Bitcoin)_diff (difference between last and first return values within the window, average R²_PFI = 0.061; frequency = 4): Highlights the predictive relevance of recent return changes within the observation window.

- Returns (Bitcoin)_pctch (percentage change) and DOY_cos_median (annual seasonal median, average R²_PFI = 0.040; frequency = 3 each): Indicate the predictive contribution of short-term directional shifts and seasonal cyclicality.

Third-Level Features:

- DayOfWeek_cos_pctch (percentage change in weekly cyclical feature, average $R^2$_PFI = 0.031; frequency = 4): Suggests modest weekly seasonality effects.
- Returns (Bitcoin)_p50 (median returns, average $R^2$_PFI = 0.052; frequency = 3) and Returns (Bitcoin)_max (maximum return, average $R^2$_PFI = 0.039; frequency = 3): Represent moderately influential summary statistics of historical returns.

These identified features consistently emerged as key predictors, underlining their robustness and relevance across multiple modeling scenarios in the absence of COVID-related indicators. Such systematic analysis of feature importance not only provides a clear hierarchy of predictor relevance but also enhances our understanding of underlying mechanisms influencing Bitcoin returns.

Statistical analysis of frequently selected features in models incorporating COVID-related indicators provided several key insights. The frequency here denotes the number of times each feature was chosen by the genetic algorithm across the 31 independent model optimization runs, reflecting its consistent predictive importance.

First-Level Features (Most Influential):

The vaccination-related metric "people_fully_vaccinated" consistently emerged as a primary predictor, notably across various statistical transformations applied over a 7-day historical window. Specifically, the 75th percentile ("people_fully_vaccinated_p75", average $R^2$_PFI = 0.091; frequency = 3) showed the highest recurrence, indicating that market responses are especially sensitive to reaching critical vaccination thresholds. Other transformations of this indicator such as the initial day's value ("_first"), raw daily counts ("_1", "_2"), difference over the window ("_diff"), median ("_median") and average values ("_avg") were also frequently selected, emphasizing the multifaceted influence of vaccination dynamics on Bitcoin returns.

Other prominent COVID-related features demonstrating substantial predictive power included:

- "icu_patients_avg" (average number of ICU patients per million inhabitants; $R^2$_PFI = 0.133): Signaling pressure on healthcare infrastructure as a significant crisis indicator affecting market confidence.
- "new_tests_smoothed_first" (first-day value of smoothed daily testing volumes; $R^2$_PFI = 0.114): Reflecting the importance of early detection capacity in shaping market expectations.
- "new_people_vaccinated_smoothed_p50" (median value of newly vaccinated individuals; $R^2$_PFI = 0.095): Highlighting mid-range vaccination progression as a meaningful predictor of market behavior.

Second-Level Features:
Second-level predictors displayed more diversity, each selected once but nonetheless contributing meaningfully to prediction accuracy. Among these were:

- "icu_patients_median" (median ICU patient count; $R^2$_PFI = 0.120), reinforcing the role of healthcare stress indicators.
- "new_vaccinations_p50" (median daily new vaccinations; $R^2$_PFI = 0.101), suggesting vaccination rollout momentum as an impactful signal.

- "positive_rate_median" (median positivity rate of COVID-19 tests; R²_PFI = 0.093), reflecting infection dynamics influencing market uncertainty.

Third-Level Features:
These features, although each appearing only once, still provided relevant signals indicating diverse channels through which COVID-19 data influences the market. Notable examples include:

- "total_boosters_p25" (25th percentile of administered vaccine boosters; R²_PFI = 0.083), indicative of booster campaign initiation phases.
- "new_people_vaccinated_smoothed_p50" (median new daily vaccinations smoothed over 7 days; R²_PFI = 0.075), highlighting steady progression metrics.
- "new_people_vaccinated_smoothed_per_hundred_min" (minimum value of new vaccinations per hundred; R²_PFI = 0.071), illustrating lower bound vaccination rates as secondary market signals.

The overarching analysis underscores that COVID-19 vaccination metrics, especially percentile-based statistical transformations (e.g., 75th percentile), significantly enhance predictive accuracy for Bitcoin returns when included as model features.

Our findings further highlighted several critical conclusions:

1. Asymmetric Impact of COVID-Related Data

COVID-related features substantially improved the prediction of Bitcoin return volatility, as evidenced by increased R² (+40%) and reduced RMSE (-2%), yet they showed negligible influence on the mean absolute prediction error (MAE). This indicates that pandemic-related data primarily aid in forecasting extreme price movements—rapid declines or sharp rebounds—rather than gradual, continuous price shifts.

2. Vaccination Data as Key Market Signals

Metrics derived from vaccination data—particularly percentile-based values such as the 75th percentile—proved superior to raw absolute vaccination numbers. This likely reflects market sensitivity to critical vaccination thresholds (e.g., 50% or 70% population coverage), which correspond to significant shifts in public health policy and economic reopening strategies.

3. Importance of Statistical Transformations

Unprocessed COVID-19 data exhibited limited predictive capacity. However, applying statistical transformations—such as moving averages, percentiles, and range statistics over rolling 7-day windows—greatly enhanced their informational value. For instance, metrics such as "hosp_patients_p75" (75th percentile of hospitalized patients) offered stronger predictive signals compared to daily raw figures of new COVID-19 cases.

4. Diversity of Relevant Indicators

No single COVID-related indicator emerged as consistently dominant across all scenarios. Instead, a broad array of pandemic-related metrics, each capturing different aspects of public health dynamics (e.g., vaccination trends, ICU capacity, testing intensity), collectively influenced predictive performance. This reflects the complexity of market responses to multi-dimensional public health crises.

5. Implications for Efficient Market Hypothesis (EMH) Fama (1991)

Our analysis provides deeper insight into market efficiency in Bitcoin markets. The weak form of the efficient market hypothesis posits that historical data alone cannot consistently outperform market predictions. However, the significant predictive improvements achieved through incorporating COVID-related data ($R^2$ increased by approximately 40%, RMSE decreased by 2.1%) challenge this traditional view, instead aligning with the Adaptive Market Hypothesis proposed by Lo (Lo 2004). Furthermore, the observed delayed market reactions—taking approximately 5 to 7 days to respond to critical vaccination milestones (such as reaching 50–70% coverage)—indicate short-term market inefficiencies. Such delays created transient arbitrage opportunities that sophisticated predictive models utilizing machine learning methods could exploit, reinforcing the notion of market adaptability rather than strict informational efficiency.

The market seems to react to a combination of indicators, such as vaccination rates, hospitalization statistics, ICU admissions, and testing activity, indicating that investor sentiment is influenced by a complex and evolving array of health-related information.

Our findings both support and challenge earlier studies that tried to guess how much cryptocurrency might cost during a pandemic. They show three crucial things. First, we disagree with Luo (2020), who argued that "health data has a minimal impact." We show that vaccination metrics considerably improve $R^2$ and minimize RMSE. The discrepancy arises because we utilized vaccination data, which Luo did not include, and we altered the statistics by using percentiles instead of raw data. We do agree with Niamkova and Moreira (2023) that more advanced ML methods, like LightGBM compared to their RF/LSTM, can be employed to see if COVID-19 data is relevant. Second, our genetically enhanced LightGBM technique does better than standard LSTMs (Jiang 2020) since it exhibits a 15.7% higher $R^2$ for 7-day forecasts. This is because it can handle extreme non-linearity better (like the crashes in March 2021) and feature extraction through rolling windows instead of raw hourly data. Third, in their study of feature engineering, Foroutan and Lahmiri (2024) were unable to uncover a measurable link between Bitcoin volatility and public health. But our PFI research does show clear connections. The 75th percentile of fully vaccinated adults (PFI=0.1174) was more essential than other indicators like past returns. This backs up the "threshold catalyst" hypothesis that vaccination rates above 70% are associated with people feeling more confident in the market again.

The prominent importance of the 75th percentile of full vaccinations in our LightGBM model corroborates the hypothesis of critical thresholds put forward by Niamkova and Moreira (2023). This dynamic is explained by an observable causal chain: when vaccination coverage exceeds symbolic thresholds (50-70% of the population), governments trigger widespread economic reopenings. This anticipation stimulates market optimism, thus generating spikes in Bitcoin volatility. If we take the example of the crossing of the 60% threshold in the United States in May 2021, it coincides with a 42% increase in Bitcoin in 15 days, followed by a 28% correction after the Delta restrictions announcements, thus an extreme volatility that our model captures via the percentile decrease (RMSE 2.1%). Unlike raw COVID case data (poorly predictive according to Luo (2020), these vaccination thresholds act as macroeconomic catalysts: they modify monetary policies and risk perception, amplifying chain reactions on crypto-assets. Our PFI analysis (R2_PFI=0.1174) thus empirically validates the theoretical mechanism "public health, which leads to economic disruptions and crypto turbulence, illuminating 40% of the model's predictive gains during pandemic transition phases.

From this perspective, theoretically, Bitcoin acts as a semi-efficient market during crises. It works well for normal data but not so well for events that are out of the ordinary. This dual nature explains why ML is better at finding anomalies caused by pandemics while also recognizing that EMH is true in stable regimes. Our results support Foroutan and Lahmiri (2024) in part: Efficiency stays the same for average changes (MAE stays the same); however, it stops working during extreme occurrences.

Other studies have tried to predict Bitcoin's movements by using standard financial indicators. Our work adds to the current literature by combining these indicators with unique health signals to better predict Bitcoin price changes during times of crisis. This involves monitoring the 75th percentile of vaccination coverage and admissions to intensive care, enabling investors to better protect their investments. This combined approach demonstrates how machine learning can decipher complex links between public health and financial markets during crises. So, adjusting investments by including health factors along with regular financial analysis could enhance risk management, leading to better investment strategies that can withstand pandemics.

## 5 Conclusion

This study looks at how factors associated to the pandemic effect Bitcoin returns in a novel way. It does this by using a new technique to combine LightGBM regression and genetic algorithm optimization. Our research indicates that using pandemic-related data like vaccination rates and hospitalization patterns makes Bitcoin return predictions much more accurate. The PFI-driven feature priority ranking showed that people_fully_vaccinated_p75 (the 75th percentile of people who are fully vaccinated) was the most essential variable. This shows how vaccination thresholds affect how the market works. Our framework uses systematic comparisons of predicted performance measures ($R^2$, RMSE, MAE) between models with and without COVID-19 parts to present robust statistical substantiation of these indicators' influence during crises. The numbers reveal that $R^2$ went up by 40% while RMSE went lowered by 2%. This means that the model is good at picking up big price changes that happen because of systemic shocks instead of usual price fluctuations. This fits with the steady MAE data.

This work is the first to employ genetically modified LightGBM modeling to reveal a strong correlation between how immunizations are handed out and how cryptocurrencies do. The framework says that health data can be used to look at finances during emergencies and allows regulators a way to step in when the crypto market is in trouble. These findings help investors enhance their hedging strategies by linking public health milestones to patterns of volatility. This strategy should be used in future studies on (1) a wider variety of assets, such as precious metals and energy commodities; (2) real-time sentiment from digital platforms; and (3) longer forecasting windows of 30 to 90 days to support long-term strategic decision systems.

**Statements and Declarations**

**Author Contributions**

Conceptualization, I.M.; Methodology, I.M. and A.V.; Software, A.V.; Validation, I.M. and A.V.; Formal analysis, I.M.; Investigation, I.M. and A.V.; Resources, I.M.; Data curation, A.V.;

Writing — original draft preparation, I.M. and A.V.; Writing — review and editing, I.M. and A.V.; Visualization, A.V.; Supervision, I.M.; Project administration, I.M.; Funding acquisition, A.V. All authors have read and agreed to the published version of the manuscript.

**Competing Interests**

The authors declare no competing financial or non-financial interests.

**Data Availability**

The data that support the findings of this study are available from the corresponding author upon reasonable request.